\documentclass[10pt,twocolumn,letterpaper]{article}

\usepackage{cvpr}
\usepackage{times}
\usepackage{epsfig}
\usepackage{graphicx}
\usepackage{amsmath}
\usepackage{amssymb}
\usepackage{enumitem}
\usepackage{amsmath}
\usepackage{eso-pic}
\usepackage{booktabs} 
\usepackage{tcolorbox} 

\newcommand{\infobox}[1]{
  \begin{tcolorbox}[
    colback=gray!7!white,  
    colframe=gray!75!black, 
    boxrule=0.75pt,  
    sharp corners,  
    top=2mm,  
    bottom=2mm,  
    left=2mm,  
    right=2mm,  
    boxsep=0.2mm,  
    ]
    #1
  \end{tcolorbox}
}

\usepackage{listings}
\usepackage{xcolor}

\definecolor{codegreen}{rgb}{0,0.6,0}
\definecolor{codegray}{rgb}{0.5,0.5,0.5}
\definecolor{codepurple}{rgb}{0.58,0,0.82}
\definecolor{backcolour}{rgb}{0.95,0.95,0.92}

\lstdefinestyle{mystyle}{
    backgroundcolor=\color{backcolour},   
    commentstyle=\color{codegreen},
    keywordstyle=\color{magenta},
    numberstyle=\tiny\color{codegray},
    stringstyle=\color{codepurple},
    basicstyle=\ttfamily\footnotesize,
    breakatwhitespace=false,         
    breaklines=true,                 
    captionpos=b,                    
    keepspaces=true,                 
    numbers=left,                    
    numbersep=5pt,                  
    showspaces=false,                
    showstringspaces=false,
    showtabs=false,                  
    tabsize=2
}

\usepackage[pagebackref=true,breaklinks=true,colorlinks,bookmarks=false]{hyperref}

\cvprfinalcopy 

\def\cvprPaperID{****} 

\ifcvprfinal\fi
\begin{document}

\title{Parameter-Efficient Fine-Tuning With Adapters}

\author{Keyu Chen
\qquad Yuan Pang
\qquad Zi Yang\\
Georgia Institute of Technology\\
{\tt\small \{kchen637, ypang43, zyang705\}@gatech.edu}}


\maketitle

\begin{abstract}
In the arena of language model fine-tuning, the traditional approaches, such as Domain-Adaptive Pretraining (DAPT) and Task-Adaptive Pretraining (TAPT), although effective, but computational intensive. This research introduces a novel adaptation method utilizing the UniPELT framework as a base and added a PromptTuning Layer, which significantly reduces the number of trainable parameters while maintaining competitive performance across various benchmarks. Our method employs adapters, which enable efficient transfer of pretrained models to new tasks with minimal retraining of the base model parameters. We evaluate our approach using three diverse datasets: the GLUE benchmark, a domain-specific dataset comprising four distinct areas, and the Stanford Question Answering Dataset 1.1 (SQuAD). Our results demonstrate that our customized adapter-based method achieves performance comparable to full model fine-tuning, DAPT+TAPT and UniPELT strategies while requiring fewer or equivalent amount of parameters. This parameter efficiency not only alleviates the computational burden but also expedites the adaptation process. The study underlines the potential of adapters in achieving high performance with significantly reduced resource consumption, suggesting a promising direction for future research in parameter-efficient fine-tuning.
\end{abstract}

\section{Introduction/Background/Motivation}

The existing literature indicates that continued pretraining on domain-specific unlabeled data can enhance the performance of language models on domain-specific tasks. Gururangan et al. (2020)\cite{Gururangan2020DontStop} demonstrate that Domain-Adaptive Pretraining (DAPT) and Task-Adaptive Pretraining (TAPT) significantly improve language models' effectiveness in the areas of biomedical, computer science, news, and product reviews. However, DAPT is resource-intensive, requiring significant computational power and data storage. The training process involves not only processing substantial amounts of domain-specific data but also adjusting the entire model's parameters to adapt to new domains. Therefore, this naturally leads us to question whether it is possible to achieve results comparable to those of DAPT and TAPT while reducing the number of parameters trained. 

To address this question, we introduce the adapter-based tuning method, proposed by Houlsby et al.(2019)\cite{Houlsby2019}. Houlsby et al.(2019) demonstrate that by using adapters, a model can be effectively transferred from one task to another while keeping most of the pretrained parameters unchanged, thus saving significant computational resources and training time. Motivated by their strategy, we employed this method to investigate whether adapter-based fine-tuning can achieve performance comparable to that of DAPT and TAPT while saving the resource.

Instead of using a single parameter-efficient language model tuning (PELT) method, we follow a new adapter method that adopts the UniPELT framework proposed by Mao et al. (2021). We employ three datasets in our study, as described in Table~\ref{tab:table1}: 1) the GLUE benchmark\cite{Wang2019GLUE}, which consists of eight different tasks designed to evaluate the general language understanding capabilities of models; 2) a domain-specific dataset tailored to our particular research focus, as described by Gururangan et al. (2020)  \cite{Gururangan2020DontStop} in their study on adapting language models to domains and tasks. This domain-specific dataset includes data from four distinct domains: Biomedicine (Biomed), Computer Science (CS), News, and Reviews, each chosen to test the adaptability and specificity of language models across varied content; 3) the Stanford Question Answering Dataset (SQuAD)\cite{DBLP}, which is a well-known benchmark for machine comprehension of text that contains 100,000+ question-answer pairs on 500+ articles. We compare the results of using adapters against those achieved through fine tuning. Our findings reveal that adapters can achieve a performance comparable fine tuning while requiring fewer trained parameters. 

Additionally, we aim to investigate whether the structure of different adapters affects the training performance of the model. We modify the UniPELT framework in three different ways: 1) We add Prompt Tuning on top of UniPELT to study whether stacking different adapters can improve performance, and whether different adapters can enhance its ability to capture more features of the data; 2) We alter the internal structure of UniPELT by replacing the first layer's LoRa with IA3 \cite{liu2022fewshot}. This method aims to study reducing the number of parameters while achieving comparable performance; 3) We stack three layers of UniPELT to observe if increasing the number of UniPELT can further enhance performance. We find that the first approach can further improve performance without significantly increasing the number of parameters involved in training.

Our project contributes in several key aspects:
\begin{itemize}[topsep=0pt, itemsep=0pt, parsep=0pt]
    \item We test the effectiveness of the UniPELT framework on domain-specific data and find that UniPELT delivers strong performance in these specialized areas. Moreover, this method significantly reduces the number of parameters that need to be trained, thereby lessening the computational resources required.
    \item We further explore the potential of enhancing model performance under the UniPELT framework by modifying adapters. This approach proves effective across most datasets in GLUE and for most of the domain-specific data, which merits further research.
    \item We apply UniPELT to the Stanford Question Answering Dataset (SQuAD) and observe positive results, demonstrating its efficacy for question-and-answer data.
\end{itemize}

\begin{table*}[ht]
\centering
\small 
\label{tab:task-descriptions}
\begin{tabular}{@{}ccccccc@{}}
\toprule
\hline
\textbf{Corpus} & \textbf{Train} & \textbf{Test} & \textbf{Task} & \textbf{Metrics} & \textbf{Domain} & \textbf{Classes} \\ \hline
\multicolumn{7}{c}{\textbf{Single-Sentence Tasks}} \\ \hline
CoLA & 8.5k & 1k & acceptability & Matthews corr. & misc. & 2 \\
SST-2 & 67k & 0.87k & sentiment & acc. & movie reviews & 2 \\ \hline
\multicolumn{7}{c}{\textbf{Similarity and Paraphrase Tasks}} \\ \hline
MRPC & 3.7k & 0.4k & paraphrase & F1 & news & 2 \\
STS-B & 5.7k & 1.5k & sentence similarity & Spearman corr. & misc. & Continuous (0-5) \\
QQP & 364k & 9.8k & paraphrase & F1 & social QA questions & 2 \\ \hline
\multicolumn{7}{c}{\textbf{Inference Tasks}} \\ \hline
MNLI & 393k & 20k & NLI & matched acc. & misc. & 3 \\
QNLI & 105k & 5.4k & QA/NLI & acc. & Wikipedia & 2 \\
RTE & 2.5k & 0.3k & NLI & acc. & news, Wikipedia & 2 \\ \hline
\multicolumn{7}{c}{\textbf{Domain Tasks}} \\ \hline
CHEMPROT & 4.1k & 3.5k & relation classification & micro-F1 & BioMed & 13 \\
RCT & 180k & 30k & abstract sent. roles & micro-F1 & BioMed & 5 \\
ACL-ARC & 1.7k & 139 & citation intent & macro-F1 & CS & 6 \\
SCIERC & 3.2k & 1k & relation classification & macro-F1 & CS & 7 \\
HYPERPARTISAN & 0.5k & 65 & partisanship & macro-F1 & News & 2 \\
AGNEWS & 115k & 7.6k & topic & macro-F1 & News & 4 \\
HELPFULNESS & 115.2k & 25k & review helpfulness & macro-F1 & Reviews & 2 \\
IMDB & 20k & 25k & review sentiment & macro-F1 & Reviews & 2 \\ \hline
\multicolumn{7}{c}{\textbf{Question and Answering}} \\ \hline
SQUAD1.1 & 87k & 10.5k & QA & f1/exact match & QA, Wikipedia & answer span \\
\bottomrule
\end{tabular}

\caption{Task descriptions and statistics.}
\label{tab:table1}
\end{table*}


\section{Approach}

\begin{itemize}
    \setlength\itemsep{-0.5em} 
    \item Batch size: 16
    \item Learning rate: \(5 \times 10^{-4}\), \(2 \times 10^{-4}\)
    \item Number of epochs: 50 with early stopping (a patience of 10 non-increasing epochs)
    \item Dropout probability: 0.1
    \item Loss function: Classification task - Cross-Entropy
\end{itemize}

In this project, we utilized the Adapters library and the Huggingface Transformers framework to facilitate our research and development efforts\cite{Poth2023Adapters,HuggingFaceQACourse}.

Our initial step involved exploring the adapter space to identify a suitable baseline adapter that demonstrated both intriguing characteristics and robust performance. We selected the UNIEPLT adapter for this purpose. Subsequently, we chose a model for adaptation based on several critical criteria.

We opted for the RoBERTa-Base model due to the following reasons:

\begin{itemize}
    \small 
    \setlength\itemsep{-0.5em} 
    \item Existing fine-tuned versions of this model are available for all tasks we intended to explore, including GLUE tasks, domain-specific tasks, and question answering.
    \item There are no existing adaptations of the RoBERTa-Base model for the UNIEPLT adapter in the Huggingface model hub, indicating an opportunity for novel contributions.
    \item The model has not yet been adapted for specific domain datasets we aimed to investigate, such as BIOMED, CS, NEWS, and REVIEWS.
\end{itemize}
\normalsize 

For the adaptation process, we fully adopted UNIEPLT's hyper-parameter tuning strategy. Specifically, we set the batch size to 16 and the input length to 128, with training conducted over 50 epochs. We incorporated early stopping based on a patience threshold of 10 epochs, where no improvement in performance metrics was observed. The learning rates were set to $2 \times 10^{-4}$ and $5 \times 10^{-4}$, respectively.

Finally, we implemented the UNIEPLT adapter on the RoBERTa-Base model and made further architectural improvements to enhance its performance and applicability to our targeted tasks by:

\begin{itemize}
    \small 
    \setlength\itemsep{-0.5em} 
    \item PT + UniPELT
    \item IA3 + Prefix + SeqBn
    \item UniPELT(Stack 3)
\end{itemize}

\begin{lstlisting}[
  language=Python,
  caption={Our PT+UniPELT added an additional PrefixTuningConfig on the original UniPELT adapter, where it ``combines LoRA, Prefix Tuning, and bottleneck adapters in a single unified setup. It additionally introduces a gating mechanism that controls the activation of the different adapter modules."\cite{UniPELTissue674}},
  basicstyle=\ttfamily\scriptsize,  
  xleftmargin=-2mm,  % Reduce the left margin
  xrightmargin=-13mm  % Reduce the right margin
]
config_adapter = ConfigUnion(
    "PromptTuningConfig(prompt_length=10),"
    LoRAConfig(r=8, use_gating=True),
    PrefixTuningConfig(prefix_length=10, use_gating=True),
    SeqBnConfig(reduction_factor=16, use_gating=True))
model.add_adapter(adapter_name, config=config_adapter)
\end{lstlisting}

\begin{lstlisting}[
  language=Python,
  caption={Our IA3+Prefix+SeqBn.},
  basicstyle=\ttfamily\scriptsize,  
  xleftmargin=-2mm,  % Reduce the left margin
  xrightmargin=-8mm  % Reduce the right margin
]
config_adapter = ConfigUnion(
    IA3Config(use_gating=True),
    PrefixTuningConfig(prefix_length=10, use_gating=True),
    SeqBnConfig(reduction_factor=16, use_gating=True))
\end{lstlisting}

\begin{lstlisting}[
  language=Python,
  caption={Our UniPELT STACK},
  basicstyle=\ttfamily\scriptsize,  
  xleftmargin=-2mm,  % Reduce the left margin
  xrightmargin=-6mm  % Reduce the right margin
]
config_adapter = ConfigUnion(
    LoRAConfig(alpha = 2, r=8, use_gating=True),
    PrefixTuningConfig(prefix_length=10, use_gating=True),
    SeqBnConfig(reduction_factor=16, use_gating=True))
model.add_adapter("unipelt1", config=config_adapter)
model.add_adapter("unipelt2", config=config_adapter)
model.add_adapter("unipelt3", config=config_adapter)
model.active_adapters = ac.Stack("unipelt1", "unipelt2", "unipelt3")
\end{lstlisting}

One significant challenge we anticipated was the potential difficulty in surpassing the UNIEPLT adapter's well-established benchmark results proposed by Facebook/Meta. Initially, our attempts did indeed under perform; however, subsequent experiments fortuitously led to solutions that outperformed UNIEPLT in tasks related to the GLUE benchmark and domain-specific adaptation.

Unexpectedly, during our implementation phase, we discovered discrepancies in the UNIEPLT adapter's code within the Adapters library. Contrary to the specifications in the original publication, which stated that the LoRA layer within UNIEPLT should have an alpha parameter of 2, the implementation used an alpha of 8. This critical issue was identified through our scrutiny of the source code and subsequently reported to the maintainers via a GitHub issue. The library's developers have since acknowledged and corrected this discrepancy in their implementation \cite{UniPELTissue674}.

\section{Experiments}

\infobox{
    \textbf{Fine-tuning (roberta-base))} \quad As our baseline, we took the fine-tuned model results from Liu et al. (2019)/cite{Liu2019RoBERTa} and Gururangan et al. (2020)
}

\infobox{
    \textbf{UniPELT (Adapter Library)} \quad As our baseline, we use an off-the-shelf RoBERTa-base model and added UniPELT to supervised fine-tuning of its parameters for each task. UniPELT is a unified framework consisting of LoRA, Prefix, and SeqBn submodules, with learnable gating parameters to control the activations of each submodule.
}

\infobox{
    \textbf{UniPELT (Paper)} \quad UniPELT( Adapter Library) is the default inplementation from Adapters library on UniPELT, which after our investgation on it's source code we find out it is inccorect, due to it set alpha = 8 in Lora. UNIEPLT (Paper) in our table is the correct implementation on UniPELT where we correct the alpha from 8 to 2 to match the orgiinal paper. 
}

\infobox{
    \textbf{PT + UniPELT(Adapter Library/ Paper)} \quad UniPELT framework with additional stacking of the prompt tuning adapter.
}

\infobox{
    \textbf{IA3 + Prefix + SeqBn} \quad Similar to the UniPELT framework, this adapter composition replaces the LoRA adapter with the IA3 in the architecture, aiming to further reduce the trainable parameters.
}

\section{Results and Discussion}

\begin{table*}[!htbp]
\small 
  \centering
  \begin{tabular}{lccccccccc}
    \toprule
    \textbf{Method} & \textbf{SST-2} & \textbf{MRPC} & \textbf{CoLA} & \textbf{RTE} & \textbf{QNLI} & \textbf{STS-B} & \textbf{MNLI} & \textbf{QQP} & \textbf{Avg.} \\
\hline
    \midrule
    Fine-tuning(roberta-base) & 94.80 & 90.20 & 63.60 & 78.00 & 92.80 & 91.20 & 87.60 & 91.90 & 86.35 \\
\hline
    UniPELT (Adapter Library) & 94.27 & \underline{89.90} & 64.45 & 71.84 & 92.71 & 90.51 & 87.21 & \underline{90.28} & 85.15 \\
    UniPELT (Paper) & \textbf{94.84}  & 87.10 & 64.36 & \textbf{77.61} & 92.53 & \underline{90.73} & \textbf{87.48} & \underline{90.28} &  \underline{85.62} \\
    UniPELT (Stack 3) & \textbf{94.84} & 88.83 &  \underline{64.92} & 73.65 & 91.96 & 90.29 & 86.25 & \textbf{90.36} & 85.15 \\
    PT + UniPELT (Adapter Library) & \underline{94.72}  &  \textbf{90.90} & \textbf{66.14} & 72.56 & \underline{92.77} & 90.60 & 87.31 & \underline{90.28} &  \textbf{85.66} \\
    PT + UniPELT (Paper) & 94.61 & 87.92 & 64.80 & \underline{74.73}  & \textbf{93.12} &\textbf{90.89} & \underline{87.46} & 90.04 & 85.45 \\
    IA3+Prefix+SeqBn & 94.15 & 78.81 & 62.37 & 58.12 & 92.44 & 88.66 & 86.79 & 90.25 & 81.46 \\
\hline
    \bottomrule
  \end{tabular}
  \caption{Results on the GLUE benchmark when all training samples are used.  We follow the same evaluation and hyper-parameter strategy from the original UniPELT Paper\cite{Mao2022UniPELT} which we used ``Matthew’s Correlation for CoLA, F1 for MRPC and QQP, Spearman’s correlation for STS-B, and accuracy for the rest. For MNLI, we evaluate on the matched dataset." we adopt UniPELT's official training scripts by set ``the input length to 128, training batch size to 16, epochs to 50 and adopt early stopping with a patience of 10 non-increasing epochs." Then tune their learning rates from {2e-4 and 5e-4} respectively. We report the best results. }
  \label{tab:table2}
\end{table*}

\begin{table}[!htbp]
\small 
\centering
\begin{tabular}{lcc}
\hline
\textbf{Method} & \textbf{Param} & \textbf{\%Param} \\
\hline\hline
Fine-tuning (roberta-base) & 124,645,632 & 100 \\
UniPELT (Paper/Adapter) & 11,083,376 & 8.892 \\
UniPELT (Stack 3) & 33,250,128 & 26.68 \\ 
PT + UniPELT (Paper/Adapter) &  11,091,056 &  8.898 \\
IA3 + Prefix + SeqBn & 10,852,988 & 8.707 \\
\hline
\end{tabular}
\caption{Total Parameters}
  \label{tab:table3}
\end{table}

\begin{table*}[!htbp]
\small 
  \centering
  \label{tab:domain-results}
\begin{tabular}{lcccccccc}
\hline
\textbf{Method} & \textbf{CHEM.} & \textbf{RCT} & \textbf{HYPER.} & \textbf{AG.} & \textbf{IMDB} & \textbf{HELP.} & \textbf{SCI.} & \textbf{ACL-ARC} \\
\hline
Fine Tuning & 81.9 & 87.2 & 86.6 & 93.9 & 95.0 & 65.1 & 77.3 & 63.0 \\
DAPT & \underline{84.2} & 87.6 & \textbf{88.2} & 93.9 & \textbf{95.4} & 66.5 & 80.8 & 75.4 \\
¬DAPT & 79.4 & 86.9 & 76.4 & 93.5 & \underline{94.1} & 65.1 & 79.2 & 66.4 \\
UniPELT (Library) &  83.40 & 88.00  & 79.28 & \underline{94.36} & 91.30 & \textbf{70.56} & 86.58  & \underline{76.70} \\
PT + UniPELT(Adapter) & \textbf{84.34} & \textbf{88.04} & \underline{80.31} & 94.27 & 91.60 & \underline{70.32} & \underline{86.81} & \textbf{82.10} \\
PT + UniPELT(Paper) & 83.93 & \underline{88.03} & 76.28 & \textbf{94.42} & 91.50 & 70.14 & \textbf{87.23} & \textbf{82.10} \\
IA3 + Prefix + SeqBn(Adapter) & 83.11 & 87.86 & 79.28 & 94.10 & 90.42 & \underline{70.51} & 81.02 & 67.99 \\

\hline
\end{tabular}
\caption{We applied the evaluation methodology from the Gururangan et al. (2020) across diverse domain tasks, using F1-micro score for BioMed tasks like CHEMPROT\cite{Kringelum2016} and RCT\cite{dernoncourt-lee-2017-pubmed}, and F1-macro for CS, News, and Reviews tasks like ACL-ARC\cite{jurgens-etal-2018-measuring}, SCIERC\cite{luan-etal-2018-multi}, HYPERPARTISAN\cite{kiesel-etal-2019-semeval}, HELPFULNESS\cite{McAuley2015}and IMDB\cite{maas-etal-2011-learning} methodology entails setting the input length to 128, with a training batch size of 16, running for 50 epochs.We report the highest scores achieved to demonstrate the efficacy of our approach.}
  \label{tab:table4}
\end{table*}

\begin{table}[!htbp]
\small 
\centering

\small 
\setlength{\tabcolsep}{9pt} 
\begin{tabular}{lcc}
\hline
\textbf{Method} & \textbf{F1} & \textbf{EM} \\ 
\hline\hline
Fine-tuning (roberta-base) & 94.6 & 88.9 \\
UniPELT (Adapter Lib.) & \underline{90.08} & \underline{82.19} \\ 
UniPELT (Paper) &  \textbf{90.23} & \textbf{82.37} \\
PT + UniPELT (Adapter Lib.) & \textit{88.38} & \textit{79.70} \\ 
PT + UniPELT (Paper) & 88.70 & 80.74 \\
IA3 + Prefix + SeqBn & 88.56 & 79.87 \\
\hline
\end{tabular}
\caption{Results on SQUAD1.1}
  \label{tab:table5}
\end{table}

To rigorously evaluate the performance of our model, we conduct a comparative analysis against established benchmarks by Facebook/Meta. Our evaluation encompasses a broad spectrum of datasets, including those from the GLUE benchmark, several domain-specific datasets, and the Stanford SQuAD.In details, we compare the results of our model with those from Roberta's fine-tuning and the unmodified UniPELT framework across the GLUE benchmark, selected domain-specific datasets, and the Q\&A dataset. Additionally, for these domain-specific datasets, we also compare our results with those obtained from DAPT and DAPT+TAPT. Furthermore, we assess model complexity by comparing the total number of trainable parameters. We measure our success by aiming to demonstrate that our model can achieve comparable or superior performance to established fine-tuning and DAPT(+TAPT) benchmarks, while simultaneously reducing both training time and the number of parameters involved.

\subsection{Complexity}
The complexity of the proposed adapter architecture can be quantified by the total number of trainable parameters, as shown in Table~\ref{tab:table3}. Compared to the base RoBERTa model, UniPELT comprises 8.892\% of the total trainable parameters, which include those of the LoRA, Prefix, SeqBn submodules, and the gating parameters that regulate the submodule activation. Conversely, despite substituting the LoRA adapter with the even more parameter-efficient IA3 adapter, which utilizes three learned vectors to rescale the keys and values in the self-attention and encoder-decoder attention layers, as opposed to LoRA's decomposed matrices, there was no significant reduction in trainable parameters from 8.892\% to 8.707\%. Moreover, the addition of the prompt-tuning layer resulted in a negligible increase in parameters, from 8.892\% to 8.898\%. However, stacked UniPELT significantly increased the total number of trainable parameters as this approach scales the adapters dimensions additively. Given that all proposed adapter architectures have comparable level of complexity, the factor of computational resourse can be disregarded when assessing the overall architectures effectiveness. Therefore, the choice between these architectures should primarily consider their performance outcomes.

\subsection{GLUE Analysis}
\subsubsection{Data}

We first conduct our experiment based on the GLUE benchmark, which is a common-used collection of datasets used to evaluate the performance of models in natural language understanding tasks. This collection of datasets can be grouped by task types: 1)Single-Sentence Tasks evaluate CoLA (8.5k train, 1k test) for grammatical acceptability with Matthews correlation, and SST-2 (67k train, 0.87k test) for sentiment analysis using accuracy.; 2)Similarity and Paraphrase Tasks include MRPC (3.7k train, 0.4k test) for paraphrase detection measured by F1 score, STS-B (5.7k train, 1.5k test) for sentence similarity with Spearman's correlation, and QQP (364k train, 9.8k test) for paraphrase questions, also evaluated with F1 score; 3)Inference Tasks feature MNLI (393k train, 20k test), QNLI (105k train, 5.4k test) and RTE (2.5k train, 0.3k test) for natural language inference, all assessed by accuracy. These tasks are designed to test models across a spectrum of language understanding from structural to inferential comprehension.

\subsubsection{Analysis}
Based on the experimental results presented In Table~\ref{tab:table2}, the method with the highest average score is Fine-tuning (roberta-base) with a score of 86.35. While the UniPELT methods and other approaches have lower average scores, their performance is still comparable to Fine-tuning (roberta-base). The performance scores across the GLUE datasets highlight the strengths of different methods. For SST-2, both UniPELT (Paper) and UniPELT (Stack 3, later) come out on top with a score of 94.84. Moving to the MRPC and CoLA datasets, PT + UniPELT (Adapter Library) takes the lead with respective scores of 90.90 and 66.14. In the RTE dataset, the highest score is secured by Fine-tuning (roberta-base) at 78.00. For QNLI, PT + UniPELT (Paper) stands out with an impressive score of 93.12. Fine-tuning (roberta-base) shows its versatility by leading in the STS-B dataset with a score of 91.20. In the MNLI matched dataset, it also achieves the top score of 87.60. Lastly, for QQP, Fine-tuning (roberta-base) continues its streak, attaining the highest score of 91.90.This suggests that although they may not exceed the fine-tuning baseline in terms of the average score, they are still competitive and their results are within a similar range, indicating effectiveness in the tasks they were tested on.

Our PT + UniPELT  method, while marginally trailing behind the Fine-tuning (roberta-base) in terms of performance, is notably efficient with respect to the number of parameters involved. PT introduces prompts that act as a steer for the model’s inference direction, minimizing the need for comprehensive retraining. Concurrently, UniPELT unites various efficient fine-tuning methodologies, including LoRA, Prefix Tuning, and Bottleneck Adapters, into a singular, controllable framework courtesy of its gating mechanism. This method uses only 11,091,056 parameters, which is just 8.898\% of the total 124,645,632 parameters that the Fine-tuning method employs. This demonstrates that our PT + UniPELT approach achieves competitive results with a significantly reduced parameter count, enhancing its appeal for scenarios where model efficiency is a priority. 

Comparing UniPELT (Stack 3) with UniPELT (Paper), we find that while the former is essentially a triple-stacked version of the latter, implying a threefold increase in the number of trainable parameters, this increase does not necessarily translate to a significant improvement in performance. In fact, for certain tasks like SST-2, MRPC, and QNLI, UniPELT (Paper) either matches or outperforms UniPELT (Stack 3) despite using fewer parameters in training. The increased computational cost due to the higher number of parameters in UniPELT (Stack 3) highlights an important consideration in the use of adapters: simply scaling up the number of trainable parameters or stacking the number of adapters doesn't guarantee better performance. This suggests that the effectiveness of an adapter is not just a function of size, but also of the structural and functional synergy within its design. Therefore, it implies that researchers may need to explore different architectural configurations to enhance performance without necessarily increasing the computational overhead.

\subsection{Domain Analysis} 
\subsubsection{Data}
In this work, we focus on the performance of adapters trained on selected domain datasets as summarized in Table~\ref{tab:table1}. These adapters are benchmarked against both fine-tuned RoBERTa and DAPT-trained RoBERTa. The domains selected for this study include BioMed, Computer Science (CS), Reviews, and News, with training set sizes varied among these datasets. Additionally, we compared the impact of vocabulary overlap across different domains on the adapter performances. Comparing with the training corpus used for training the base Roberta, the News domain has the highest vocabulary overlap of 54.1\%, followed by Reviews domain of 34.5\%. Domains such as BioMed and CS have more specialized vocabulary having relatively lower olverap of 27.3 and 19.2, respectively.    

\subsubsection{Analysis}

As demonstrated in Table~\ref{tab:table4}, the adapters, trained on selected domain datasets, generally surpassed both the fine-tuned RoBERTa and the DAPT-trained RoBERTa across most tasks, with the notable exceptions of the Hyperpartisan news and IMDb review tasks. These exceptions revealed that the adapters were less effective with reduced number of parameters compared to the base RoBERTa, particularly with small training datasets. Nonetheless, the adapters achieved performance on par with DAPT-RoBERTa and the base model, largely due to the high vocabulary overlap with RoBERTa's training corpus. Conversely, in domains such as BioMed and CS where vocabulary overlap is substantially lower, the adapters demonstrated a significant improvement in F1 scores, indicating the adapters effectively capturing nuances that were unseen to the base RoBERTa. This indicates that adapter modules can be finely tuned to boost performance in specialized tasks, even with a reduced number of learable parameters. The impact is especially noticeable in areas with distinct vocabularies, such as CS, where we noted a substantial rise in F1 scores—from 63.0 to 82.1 for ACL-ARC and from 77.3 to 86.81 for SCIERC. Furthermore, the training set sizes for ACL-ARC and SCIERC are 1,688 and 3,219, respectively—orders of magnitude smaller than those for Reviews and News. This suggests that the domain specificity of adapters takes advantage of the base RoBERTa's fundamental language processing ability while focusing on learning from specialized and distinct vocabulary, hence when giving the right data, it does not require a large dataset.However, where vocabulary overlap was high, the benefit of domain-specific training on adapters diminished. This was apparent in the Ag News dataset, where the extensive 115,000 training dataset only had a minor improvement in the F1 score from 93.9 to 94.27, suggesting that the high vocabulary overlap may have rendered the additional domain training superfluous,or in another order, the domain datasets are not ``fresh" enough, as the base RoBERTa had already learned a large portion of the relevant vocabulary through the fine-tuning. When examining the performance of BioMed domain, we observed a modest improvement in F1 scores for both CHEMPROT (from 81.9 to 84.34) and RCT datasets (from 87.2 to 88.04), despite the low vocabulary overlap and the relatively large size of the datasets  than CS. This could be hypothesized to be attributable to the high heterogeneity of domain-specific vocabulary and context in the BioMed domain. However, the training datasets may not have provided sufficient domain-specific contextual training, instead supplying nuances that RoBERTa had already learned. Coupled with the higher number of classes to predict, the adapters did not yield as significant an improvement as in the CS domain. Furthermore, this could suggest a limitation of the parameter-efficient tuning strategy, where a larger number of parameters may be necessary to tackle domains that have highly heterogeneous and complex contexts.Upon comparing the adapter architectures, we found that adding an extra adapter layer for prompt-tuning significantly outperformed the UniPELT or internal substitution within UniPELT submodules (IA3+Prefix+SeqBn), especially for the small ACL-ARC dataset in the CS domain. The PT+UniPELT achieved an F1 score of 82.10, markedly higher than IA3+Prefix+SeqBn (67.99) and UniPELT (76.70). This indicates that the additional prompt-tuning layer is highly effective in distilling the  input that enhances the adapter's ability to capture domain knowledge, considering it introduced only an additional 0.185\% in the total trainable parameters. Therefore, these results confirmed that the layer-wise stacking based on UniPELT has superior performance compared to UniPELT or UniPELT with compositional alterations.

\subsection{SQuAD}
\subsubsection{Data}
Finally, we adopt the same strategy to Houlsby, ``we conﬁrm that adapters work on tasks other than classiﬁcation by running on SQuAD v1.1 (Rajpurkar et al., 2018)." \cite{Houlsby2019,DBLP}. We explore similar themes in our work as noted in the previous research by as the Houlsby et al. (2019)\cite{Houlsby2019}. 

\subsubsection{Analysis}
In addition to studying the effectiveness of adapters in classification tasks, we also wish to explore whether this approach is effective for text generation tasks. This is particularly important in today's context of popular generative AI, as training models to generate outputs that more closely align with user needs while reducing computational resources is crucial. In our experiments on the Stanford SQuAD in Table~\ref{tab:table5}, we employ the identical hyperparameters previously used, omitting early stopping. The outcomes proved unexpected, as while the combination of Pre-training (PT) with UniPELT consistently excelled across most tasks within the GLUE benchmark and Domain Adaptation tasks, it notably underperformed on the Stanford SQuAD compared to the standalone UniPELT framework by 1.5\%. This divergence underscores the complexity of task-specific model behavior and the need for tailored approaches to model adaptation and training. However, to confirm why, more experiments is needed.

\section{Conclusion}
In this study, we have demonstrated that adapters trained with domain-specific knowledge  significantly outperformed the base RoBERTa model, without the need for  re-training of a large number of base model parameters. We evaluate the performance by using GLUE dataset. This method is especially advantageous in domains where there is minimal vocabulary overlap with the base RoBERTa model. Our results indicate that adapters can competently tackle multi-label classification tasks even with relatively small domain-specific datasets, suggesting the effective architecture of the proposed adapters frameworks. Additionally, we found that layer-wise stacking of adapters based on UniPELT proved to be more effective than altering the UniPELT's internal compositions. Furthermore, we test this method for text generation tasks by using SQuAD data and find good performance as well.





{\small
\bibliographystyle{ieee_fullname}
\bibliography{egbib}
}

\end{document}